\documentclass[conference]{IEEEtran}
\usepackage[left=1.62cm,right=1.62cm,top=1.9cm]{geometry}
\usepackage{fancyhdr}
\usepackage{amsmath,amssymb,amsfonts}
\usepackage{algorithm}
\usepackage{algpseudocode}  
\usepackage{amsmath}
\usepackage{booktabs}
\usepackage{array}
\usepackage{caption}
\usepackage{cite}
\usepackage{comment}

\usepackage[normalem]{ulem}
\usepackage{graphicx}
\usepackage{graphbox}
\usepackage{textcomp}
\usepackage{hyperref}
\usepackage{booktabs}
\usepackage{booktabs}
\usepackage{adjustbox}
\usepackage{multirow}

\usepackage{url}

\usepackage{xcolor}
\usepackage{xspace}
\usepackage{acronym}
\acrodef{rmse}[RMSE]{root mean squared error}
\acrodef{mae}[MAE]{mean absolute error}
\acrodef{5g}[5G]{fifth generation mobile communication}
\acrodef{jsd}[JSD]{jensen-shannon divergence}
\acrodef{wd}[WD]{wasserstein distance}
\acrodef{kl}[KL]{kullback-leibler}
\acrodef{rnn}[RNN]{recurrent neural network}
\acrodef{lstm}[LSTM]{long short-term memory}
\acrodef{bic}[BIC]{bayesian information criterion}
\acrodef{aic}[AIC]{akaike information criterion}
\acrodef{ar}[AR]{autoregressive}
\acrodef{ma}[MA]{moving average}
\acrodef{mlp}[MLP]{multi-layer perceptron}
\acrodef{arima}[ARIMA]{auto regressive integrated moving average}
\acrodef{sarima}[SARIMA]{seasonal auto regressive integrated moving average}
\acrodef{tkcm}[TKCM]{top-k case matching}
\acrodef{automl}[AutoML]{automated machine learning}
\acrodef{knn}[KNN]{K-nearest neighbors}
\acrodef{som}[SOM]{self organizing maps}
\acrodef{mi}[MI]{multiple imputation}
\acrodef{em}[EM]{expectation maximization}
\acrodef{ml}[ML]{machine learning}
\acrodef{dl}[DL]{deep learning}
\acrodef{iot}[IoT]{internet of things}
\acrodef{cart}[CART]{classification and regression tree}

\begin{document}

\title{Evaluation of Missing Data Imputation for Time Series Without Ground Truth}

\author{
    Rania Farjallah$^{\ast}$, Bassant Selim$^{\ast}$, Brigitte Jaumard$^{\S}$, Samr Ali$^{\dag}$, and Georges Kaddoum$^{\ddag, \star}$\\
    $^{\ast}$Systems Engineering Department, École de Technologie Supérieure, Montreal, Canada \\ 
    $^{\S}$Computer Science and Software Engineering Department (CSSE), Concordia University, Montreal, Canada \\ 
    $^{\dag}$GAIA, AI Hub Canada, Ericsson, Montreal, Canada \\ 
    $^{\ddag}$Electrical Engineering Department, École de Technologie Supérieure, Montreal, Canada \\ 
    $\star$Artificial Intelligence \& Cyber Systems Research Center, Lebanese American University, Lebanon \\
    Emails : rania.farjallah.1@ens.etsmtl.ca$^{\ast}$, bassant.selim@etsmtl.ca$^{\ast}$, brigitte.jaumard@concordia.ca$^{\S}$, \\samr.ali@ericsson.com$^{\dag}$, georges.kaddoum@etsmtl.ca$^{\ddag, \star}$
}

\maketitle
\setcounter{page}{1}
\begin{abstract}
The challenge of handling missing data in time series is critical for maintaining the accuracy and reliability of \ac{ml} models in applications like \ac{5g} network management. Traditional methods for validating imputation rely on ground truth data, which is inherently unavailable. This paper addresses this limitation by introducing two statistical metrics, the \ac{wd} and \ac{jsd}, to evaluate imputation quality without requiring ground truth. These metrics assess the alignment between the distributions of imputed and original data, providing a robust method for evaluating imputation performance based on internal structure and data consistency. We apply and test these metrics across several imputation techniques. Results demonstrate that \ac{wd} and \ac{jsd} are effective metrics for assessing the quality of missing data imputation, particularly in scenarios where ground truth data is unavailable.

Index Terms—Missing Data Imputation, Time Series Data, Machine Learning,  Wasserstein Distance, Jensen-Shannon Divergence.
\end{abstract}

\IEEEpeerreviewmaketitle

\section{Introduction}
As data traffic and application demands surge with the deployment of \ac{5g} networks, communication systems face unprecedented challenges in complexity. Applications like real-time video streaming, autonomous vehicles, and \ac{iot} devices require high data throughput, ultra-low latency, and reliable, adaptive resource management. To meet these demands, machine learning (\ac{ml}) and \ac{dl} techniques have become indispensable for optimizing network performance. These models enable adaptive traffic forecasting, real-time bandwidth allocation, efficient energy management, and proactive congestion control, which identifies and mitigates network bottlenecks to prevent performance degradation \cite{10327908, 9532728}. However, the accuracy of these \ac{ml}-driven optimizations depends heavily on high-quality, complete data—an ideal that is often unattainable, as real-world datasets are typically riddled with missing values or inconsistencies. Consequently, effective data filling strategies are essential to ensure that \ac{ml} models function reliably and unlock the full potential of \ac{5g} network capabilities. While imputation might seem similar to prediction, it differs in the availability of ground truth. Prediction forecasts future values using observed data, whereas imputation reconstructs missing values without access to ground truth.

In practice, \ac{5g} network datasets, such as performance measurement counters, frequently contain missing values due to factors like system failures, storage constraints, and synchronization issues within the network’s data collection infrastructure \cite{9808164}. These data gaps can undermine the accuracy of \ac{ml} models used for network management tasks, such as traffic forecasting and resource allocation, thereby impacting overall network performance. Traditional imputation methods, such as deletion or simple filling techniques, often fall short because they do not account for the dependencies and correlations among metrics in \ac{5g} network data. Advanced \ac{ml}-based imputation methods offer more accurate results by modeling these relationships; however, they introduce a new challenge: the need for reliable validation mechanisms to ensure that imputed values faithfully represent the missing data. Existing imputation methods typically assume access to ground truth data to validate their performance, but in data-filling tasks, this original data is inherently missing. Without ground truth, assessing the accuracy of imputed values becomes challenging, highlighting the need for alternative metrics to evaluate model performance effectively.

This paper explores the use of statistical tools as validation metrics for evaluating imputation performance. These tools measure differences or distances between distributions, a common approach in synthetic data evaluation. By comparing distributional differences between imputed data and the original data, they provide a practical method for assessing imputation accuracy without relying on ground truth. To validate the proposed metrics, we apply them to complete datasets that share characteristics with incomplete \ac{5g} network data.

The remainder of this paper is organized as follows. Section \ref{sec:Literature} describes related works. Section \ref{sec:methods} presents the proposed metrics. Section \ref{sec:methodology} illustrates the datasets and the validation methodology. Section \ref{sec:resuls} discusses the experimental settings and results, and Section \ref{sec:conlusion} concludes the paper and provides future research directions.

\section{Related Works }
\label{sec:Literature}
The quality and completeness of datasets are essential for effective analysis, especially in time series applications where missing values can obscure key patterns, such as seasonality, and reduce predictive accuracy. Seasonality, characterized by periodic fluctuations over time (e.g., daily or weekly patterns), can critically impact model performance by introducing systematic variations that imputation methods must account for. However, many imputation methods do not incorporate seasonality, relying only on remaining observed data, which simplifies modeling but reduces accuracy. 

For small gaps, basic interpolation methods are commonly used, fitting smooth curves between known data points to estimate missing values. While simple, these methods fail to capture temporal dependencies and can lead to biased results. Other approaches include single-imputation techniques, such as Hot Deck, Cold Deck, and \ac{em}, which replace each missing value with a single estimate but may not reduce bias effectively \cite{DONDERS20061087}. In contrast, \ac{mi} techniques offer advantages by providing information on how missing data impacts parameter estimates \cite{DONDERS20061087}. Advanced methods, such as regression-based imputation, \ac{som} \cite{JUNNINEN20042895}, and \ac{knn}, have proven more effective, particularly in datasets where temporal relationships are significant. \ac{knn}, for instance, fills missing values by identifying the k-closest patterns around the missing data point, utilizing local similarity to improve imputation accuracy \cite{Tarsitano2011}.

Incorporating seasonality into imputation methods can markedly improve accuracy. Techniques such as seasonal adjustment with Kalman filters and linear interpolation on seasonally decomposed data, available in tools like the forecast and zoo R-packages, have proven useful \cite{article2}. Additionally, \ac{sarima} has been applied to seasonal time series, although it struggles with consecutive missing values \cite{article}. More advanced approaches include neural network-based methods, such as \ac{mlp} \cite{atmos14020355} and \ac{lstm} networks, as well as hybrid neural models \cite{Bandara_2021}. Pattern-based methods, like the \ac{tkcm} algorithm, have also been used to handle missing data \cite{article3}. However, many of these methods are limited to single-seasonal patterns and are not well-suited for multiple seasonalities.

All imputation validation approaches rely on ground truth data to evaluate accuracy. Typically, studies create artificial gaps in complete datasets, then fill these gaps and compare the imputed values to the known originals. This approach enables traditional metrics like \ac{rmse} and \ac{mae} to measure a model’s ability to accurately reconstruct missing data. However, this validation strategy assumes that ground truth data is available—an assumption that often does not hold in real-world applications, especially in dynamic environments such as \ac{5g} networks.

This paper addresses this gap by introducing two statistical metrics to evaluate imputation methods without relying on ground truth. These metrics assess how closely the distributions of imputed data align with those of the original data, providing a way to evaluate imputation effectiveness based on internal structure and data consistency. By adapting these metrics, we offer a set of tools that complements existing validation approaches and extends evaluation capabilities to real-world applications where complete datasets are not available.

\section{Proposed Data Filling Validation Metrics}
\label{sec:methods}
To validate the performance of data imputation methods in the absence of ground truth, we propose two statistical metrics, namely the \ac{wd} and the \ac{jsd}, as described in this section, to measure distributional similarity. These metrics, originally used for synthetic data validation \cite{Stenger2024}, allow us to evaluate how closely the imputed data match the pre-gap distribution. Table \ref{tab:metric_comparison} provides a summary of the no ground truth and traditional ground truth metrics considered in this work.
\begin{table*}[h]
\centering
\caption{Comparison of no ground truth and ground truth metrics}
\label{tab:metric_comparison}
\renewcommand{\arraystretch}{1.5} 
\begin{tabular}{|p{3cm}|p{6cm}|p{6cm}|}
\hline
\textbf{Metric type} & \textbf{No ground truth metrics} & \textbf{Ground truth metrics} \\
\hline
\textbf{Metrics} & Wasserstein distance (\ac{wd}), Jensen-Shannon divergence (\ac{jsd}) & Root mean squared error (\ac{rmse}), Mean absolute error (\ac{mae}) \\
\hline
\textbf{Ground truth requirement} & Not required, uses pre-gap data as reference & Requires actual ground truth values for accuracy assessment \\
\hline
\textbf{Evaluation principle} & Evaluates distributional and structural alignment between imputed and pre-gap data & Measures direct error by comparing imputed values to true values \\

\hline
\textbf{Interpretation} & Lower values indicate better alignment with original data distribution & Lower values indicate more accurate gap filling compared to true values \\
\hline
\end{tabular}
\end{table*} 
\subsection{Wasserstein Distance}
The \ac{wd}, also known as the earth mover’s distance, measures the dissimilarity between two probability distributions by calculating the minimum effort required to transform one distribution into another \cite{Villani2009}. Formally, the \( W(P, Q) \) between two distributions \( P \) and \( Q \) over a metric space \( \mathcal{X} \) is defined as:
\begin{equation}
W(P, Q) = \inf_{\gamma \in \Gamma(P, Q)} \int_{\mathcal{X} \times \mathcal{X}} \|x - y\| \, d\gamma(x, y),
\end{equation}
\noindent where \( \Gamma(P, Q) \) represents the set of all possible joint distributions (or couplings) \( \gamma \) with marginals \( P \) and \( Q \). In this context, \( \|x - y\| \) quantifies the distance between points \( x \) and \( y \) in the space \( \mathcal{X} \). The \ac{wd} gives the minimum “cost” required to "move" probability mass from the distribution \( P \) to match distribution \( Q \), with “cost” referring to the product between the probability mass and the distance it must be moved. This metric is particularly suitable for evaluating imputed data without ground truth because it directly compares the distribution of pre-gap values with that of the imputed values. A smaller \ac{wd} means that the imputed data closely aligns with the patterns of the pre-gap values, showing that the imputation method maintains the original data characteristics well \cite{Stenger2024}. 

\subsection{Jensen-Shannon Divergence}
The Jensen-Shannon Divergence is a statistical measure for quantifying the similarity between two probability distributions \cite{Stenger2024}.  It is based on the \ac{kl} divergence, which measures the divergence of one probability distribution \( P \) from a reference distribution \( Q \). However, \ac{kl} divergence presents two limitations that make it unsuitable for evaluating imputation techniques. Firstly, \ac{kl} divergence is asymmetric, meaning that \( D_{KL}(P \| Q) \neq D_{KL}(Q \| P) \), which can lead to biased comparisons depending on the order in which the distributions are considered. Second, \ac{kl} divergence can yield infinite values if there are points in the support of \( P \) that have zero probability in \( Q \). This sensitivity to non-overlapping supports can lead to instability, producing values that are disproportionately large or undefined, even when the distributions are similar in other regions. Such behavior makes \ac{kl} divergence unreliable for assessing the similarity between pre-gap and imputed distributions. Unlike the \ac{kl} divergence, \ac{jsd} is symmetric, making it more robust for comparing distributions. The \ac{jsd} between distributions \( P \) and \( Q \) is defined as:
\begin{equation}
JS(P \| Q) = \frac{1}{2} D_{KL}(P \| M) + \frac{1}{2} D_{KL}(Q \| M)  ,
\end{equation}
\noindent where \( M = \frac{1}{2}(P + Q) \) represents the average distribution, and \( D_{KL} \) is the \ac{kl} divergence. A \ac{jsd} close to 0 indicates high similarity between distributions, signifying that the synthetic data closely replicates the real data's statistical properties. 

\section{Metrics Validation Methodology}
\label{sec:methodology}
This section outlines our approach for validating the proposed no ground truth metrics to ensure their suitability for evaluating imputation quality.

\subsection{Validation Methodology}
To ensure our approach is both effective and reliable, we introduce a validation process for the proposed metrics. By validating these metrics, we aim to confirm that, in the absence of ground truth data, they accurately reflect the performance of different imputation methods and can be considered a reliable alternative to traditional validation metrics. 

The methodology begins by validating the proposed no ground truth metrics using two complete datasets. To simulate real-world scenarios, we artificially create N gaps of different lengths at random positions within the datasets. This approach allows us to compare the gap-filled values with the true held-out observations as well as analyze and compare the performance of different gap-filling methods using both ground truth and the no ground truth metrics. The gaps are filled using three different imputation methods: an interpolation-based approach, a \ac{ml}-based approach, and a \ac{dl}-based approach.

\begin{figure}[h]
  \centering
  \includegraphics[width=0.45\textwidth]{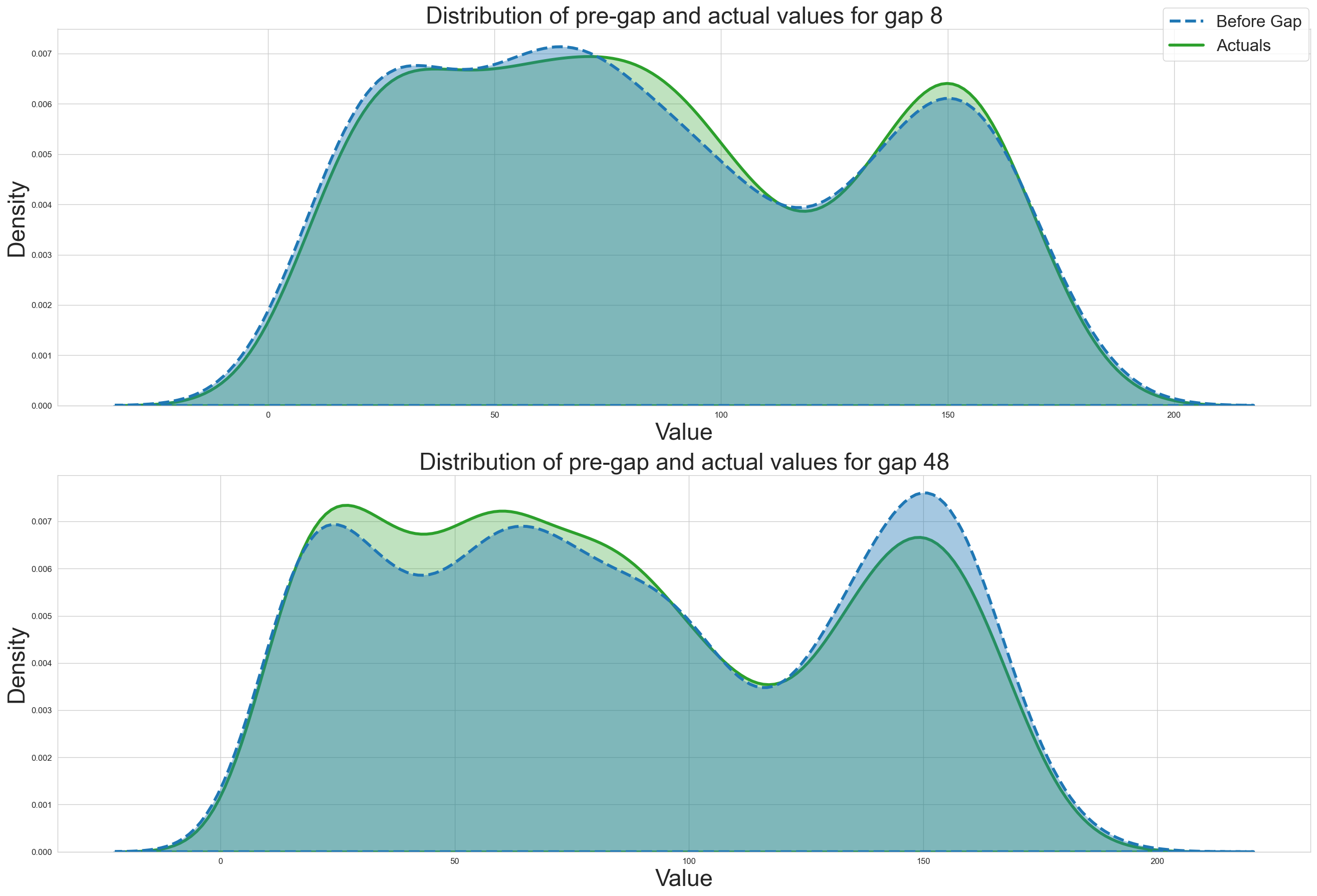}
  \caption{Distribution of the pre-gap and actual values}
  \label{fig:Image40}
\end{figure}

After filling the gaps with each method, we evaluate the accuracy of the proposed metrics by calculating and comparing them with traditional metrics, namely the \ac{rmse} and \ac{mae}, which use the original data as ground truth. For ground truth-based metrics, we compare the gap-filled values directly to the true observations. In contrast, for the proposed metrics, we use pre-gap values as a reference. The reason behind this is that we do not have ground truth data in practice, and as shown in Figure \ref{fig:Image40}, when considering the same gap size, the distribution of pre-gap values closely matches that of the true values. This similarity in distribution ensures that the pre-gap segment serves as an appropriate proxy for evaluating how well the gap-filled data aligns with the original data distribution, without needing access to true values. This approach enables us to evaluate whether the statistical metrics can effectively measure how well the gap-filled data aligns with the general data distribution, without requiring the ground truth.
 By demonstrating that the results from both types of metrics are closely aligned, we establish that the no ground truth metrics can serve as reliable alternatives for traditional metrics. 

\subsection{Description of the Datasets}
We considered two datasets to validate the proposed metrics. These datasets consist of seasonal time series data, capturing vehicle counts across various transportation modes. Since the datasets are relatively complete, they are well-suited for validating the proposed metrics against ground truth based metrics.
\subsubsection{ Telraam Dataset} This dataset tracks the movement of pedestrians, cyclists, cars, and heavy vehicles every hour during the day period. Although Telraam gathers data from various cities, we selected Brussels for its diverse traffic patterns, also due to low missing rate in this particular intersection. Data are collected with cameras installed as part of the Telraam device \cite{Telraam} recording data from October 2021 to January 2024. Data can be accessed in real-time via the Telraam API. The missing rate in this dataset is under 1\% over all the captured period.
\subsubsection{ Madrid Dataset} This dataset provides historical and real-time traffic data in the city of Madrid with a frequency of 15 min \cite{TelraamApi}. The data is recorded from July 2013 to October 2024 with a missing data of 0\%. The dataset captures the intensity of car traffic across different segments. 

\section{Numerical Results}
\label{sec:resuls}
In this section, we demonstrate that gap-filling approaches can be effectively evaluated using both ground truth and no ground truth metrics. By comparing gap-filled values to true observations and pre-gap values, we evaluate the consistency of these metrics.

\subsection{Experimental Setup}
In this study, we optimized different imputation models to accurately fill missing values in time series data. This section details the training set, model optimization techniques, and feature selection used across different imputation methods.
\subsubsection{Interpolation-Based Imputation}
Polynomial interpolation is applied directly to the available hourly data to estimate missing values. This method fits a flexible curve through known data points, ensuring continuity in the imputed data. 

\subsubsection{ARIMA-Based Imputation}
The \ac{arima} model addresses missing values by modeling both \ac{ar} and \ac{ma} components, capturing trends and autocorrelations in data. For both datasets, \ac{arima} is trained on six weeks of data. Using the \texttt{auto\_arima} function, the parameters \( p \), \( d \), and \( q \) are optimized automatically to represent the autoregressive, differencing, and moving average components of the model. 

\subsubsection{SARIMA-Based Imputation}
The \ac{sarima} model extends \ac{arima} by including seasonal terms to capture periodic patterns in data. \ac{sarima} followed the same training durations as \ac{arima}, with six weeks for both datasets, incorporating seasonal terms. The \texttt{auto\_arima} function optimizes parameters \( p \), \( d \), and \( q \) for non-seasonal components, as well as \( P \), \( D \), \( Q \), and \( s \) for seasonal terms, with \( s \) representing the seasonal frequency.

\subsubsection{XGBoost-Based Imputation}
XGBoost is an advanced boosting algorithm combining multiple weak learners to improve prediction accuracy iteratively. The model is trained on two years of data for the Madrid dataset and one year for the Telraam dataset. Additionally, we optimize four hyperparameters using GridSearch, including \texttt{max\_depth}, \texttt{learning\_rate}, \texttt{n\_estimators}, and \texttt{subsample}, to enhance model accuracy and prevent overfitting. To capture complex temporal patterns, XGBoost uses time-based features namely \textit{simple moving average (SMA)}, \textit{exponentially weighted moving average (EWMA)} and \textit{hour of day}. 

\subsubsection{LSTM-Based Imputation}
\ac{lstm} networks, a type of \ac{rnn} optimized for sequence prediction, are used to address missing values in time series data. \ac{lstm} is trained on two years of data for the Madrid dataset and one year for the Telraam dataset, with look-back periods of 20 hours for Madrid and 7 hours for Telraam. Hyperparameters including, \texttt{number of units} in the hidden layers, \texttt{dropout rate}, \texttt{learning rate}, and \texttt{batch size} are optimized using GridSearch to allow \ac{lstm} to capture temporal dependencies effectively. The \ac{lstm} model also incorporates previously mentioned time-based features used in XGBoost. 

\subsection{Results and Discussion}
In the following section, we present the performance of data filling methods using the proposed no ground truth and ground truth metrics. The models are applied to fill 100 distinct artificial gaps with lengths ranging from 2 to 48 hours. The results are then averaged over these gaps.

\begin{figure*}[h]
  \centering
  \includegraphics[width=0.95\textwidth]{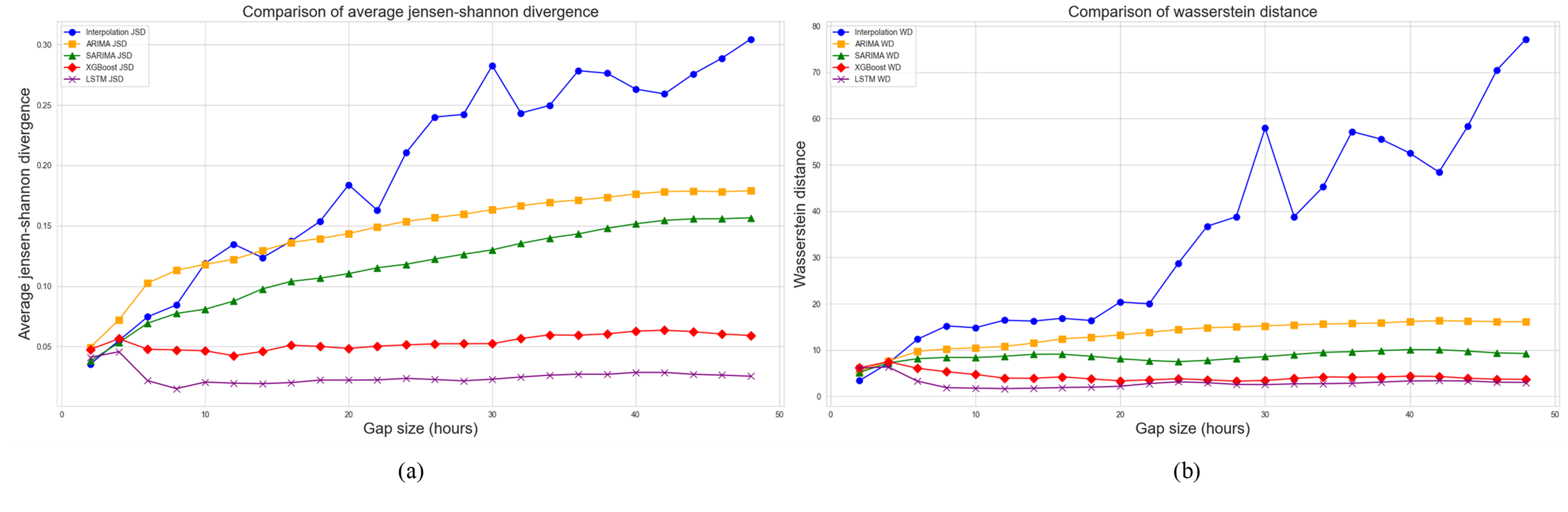}
  \caption{(a) Average Jensen-Shannon divergence and (b) Wasserstein distance for different gap sizes considering the Madrid dataset}
  \label{fig:Image31}
\end{figure*}
\begin{figure}[h]
  \centering
  \includegraphics[width=0.48\textwidth]{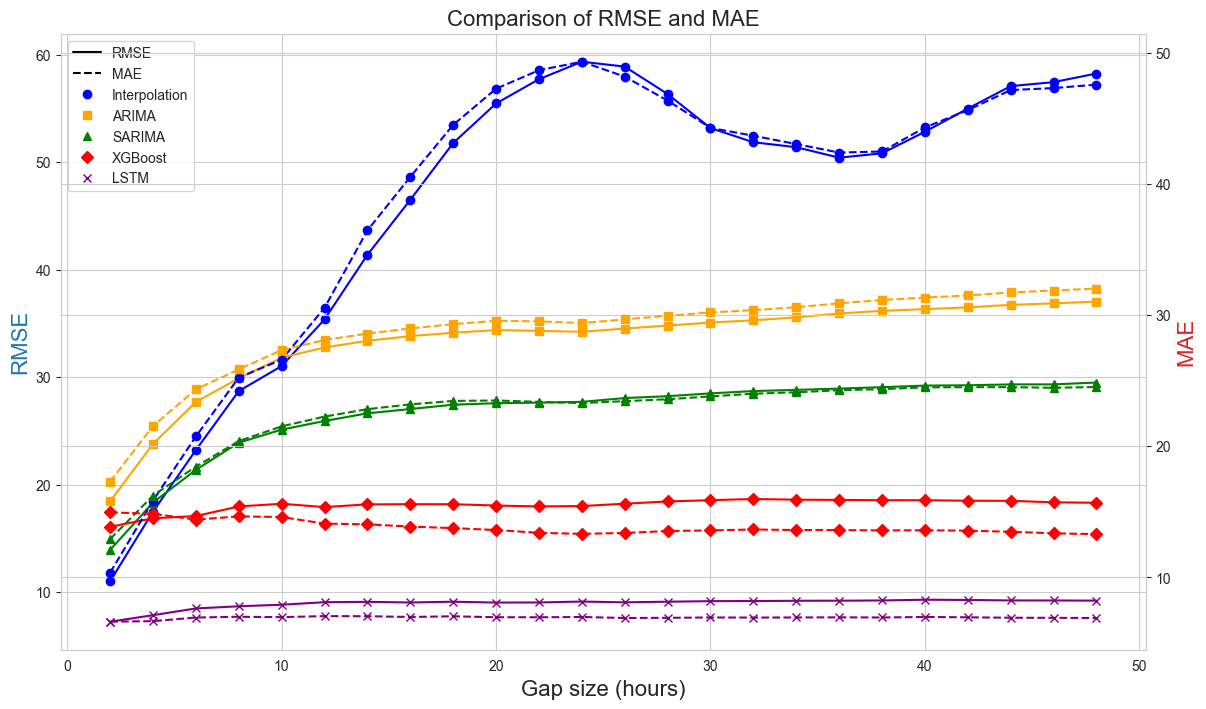}
  \caption{Results of \ac{mae} and \ac{rmse} for Madrid dataset}
  \label{fig:Image30}
\end{figure}

As shown in Figures \ref{fig:Image31} and \ref{fig:Image30}, the analysis of the Madrid dataset highlights a strong alignment between the proposed no ground truth metrics and traditional ground truth based metrics. For instance, both sets of metrics show that the \ac{lstm} model outperforms the other methods across all gap sizes. For example, in the case of a 6-hour gap, \ac{lstm} shows the lowest values for \ac{jsd} and \ac{wd}, as well as the smallest average \ac{rmse} and \ac{mae}. This similarity between the no ground truth and ground truth metrics suggests that \ac{jsd} and \ac{wd} can accurately capture the distributional consistency of the imputed values with pre gap data. Additionally, the XGBoost model also performs well in the Madrid dataset, showing a strong alignment between proposed and traditional metrics, further reinforcing that \ac{jsd} and \ac{wd} can provide insights comparable to \ac{rmse} and \ac{mae} in evaluating model effectiveness.

In contrast, Interpolation and \ac{arima} show significantly higher \ac{jsd} and \ac{wd} values, especially as gap sizes increase. This trend mirrors the increases in \ac{rmse} and \ac{mae} for these models, underscoring their limitations in preserving the original data pattern over longer gaps. These observations validate that \ac{jsd} and \ac{wd} can highlight deviations in model performance, capturing the same limitations as \ac{rmse} and \ac{mae} without needing ground truth data. Furthermore, the \ac{sarima} model performs moderately well, with lower deviations than \ac{arima} and interpolation in both \ac{jsd} and \ac{wd}. \ac{sarima} demonstrates a balanced capability to retain seasonal patterns and trends in shorter gaps, although it underperforms compared to \ac{lstm} and XGBoost for larger gaps. Overall, the close alignment between the proposed no ground truth and traditional error metrics across all gap sizes and different models confirms that the proposed no ground truth metrics can reflect model performance. The strong performance of \ac{lstm} and XGBoost underscore the performance of these models. In contrast, the higher \ac{jsd} and \ac{wd} values for Interpolation and \ac{arima} highlight their limitations, further validating the use of the proposed metrics as reliable performance metrics even in the absence of ground truth.

\begin{figure*}[h]
  \centering
  \includegraphics[width=0.95\textwidth]{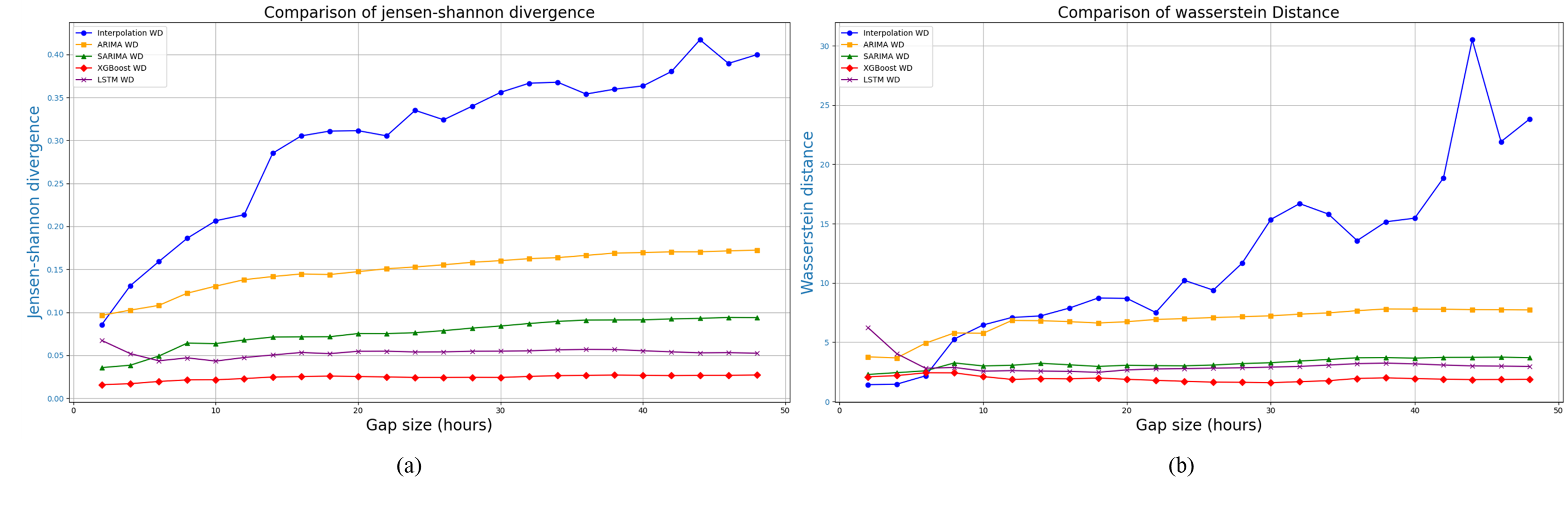}
  \caption{(a) Average Jensen-Shannon divergence and (b) Wasserstein distance for different gap sizes considering the Telram dataset}
  \label{fig:Image35}
\end{figure*}

\begin{figure}[h]
  \centering
  \includegraphics[width=0.48\textwidth]{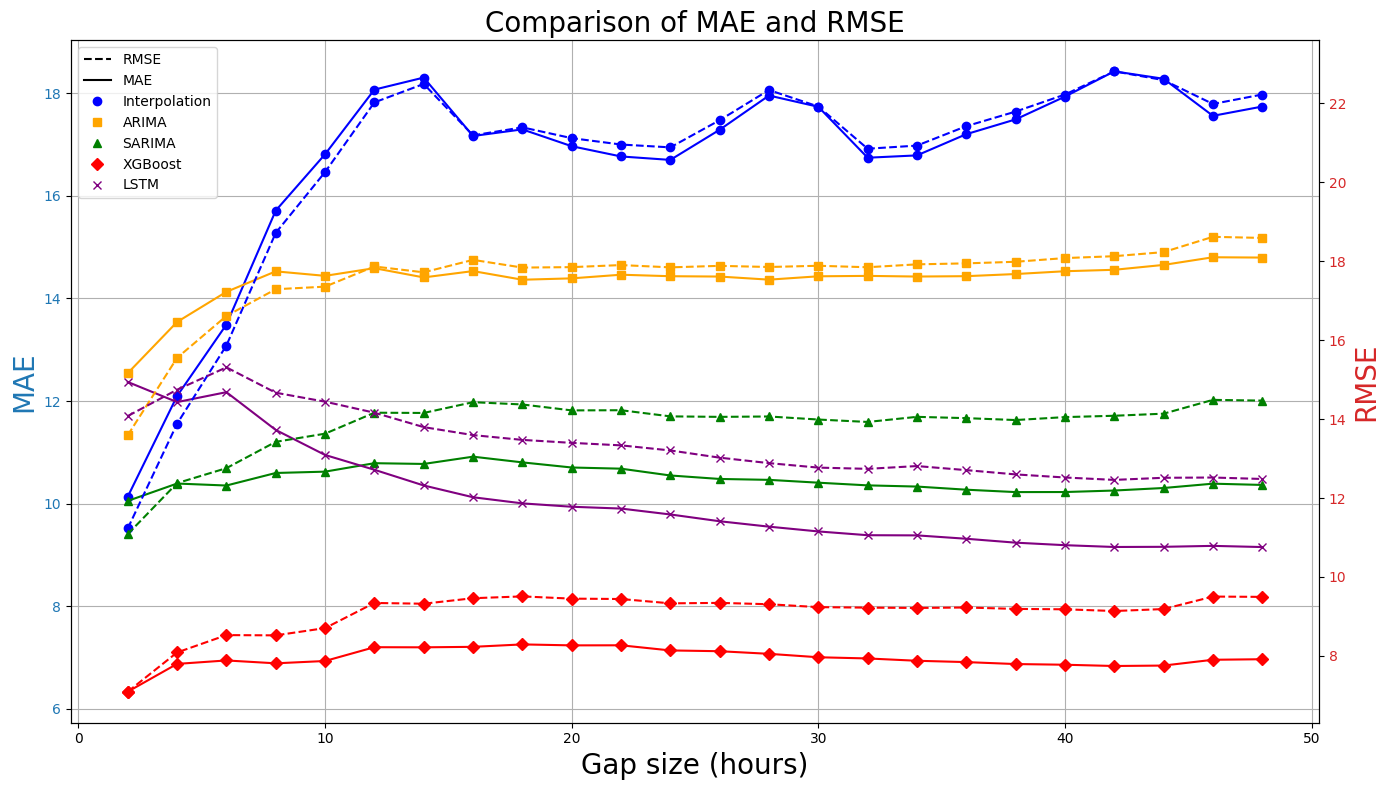}
  \caption{Results of \ac{mae} and \ac{rmse} for Telraam dataset}
  \label{fig:Image34}
\end{figure}

For the Telraam dataset, we apply the same gap-filling methods and evaluate them using the proposed no ground truth and ground truth based metrics, as illustrated in Figures \ref{fig:Image35} and \ref{fig:Image34}. Across all gap sizes, XGBoost consistently shows the lowest \ac{jsd} and \ac{wd} values, which indicates that it closely maintains the original data’s distribution, even for longer gaps. 
Furthermore, traditional ground truth based metrics also confirm XGBoost’s superior performance, particularly for larger gaps, where it sustains low error rates compared to other models. In contrast, the \ac{lstm} model also performs well on the Telraam dataset but shows higher \ac{jsd} and \ac{wd} values than XGBoost. While \ac{lstm} achieves comparable performance for shorter gaps, this divergence may be due to the limited size of the training set, which affects \ac{lstm}'s ability to generalize over larger gaps and makes it more sensitive to certain underlying patterns in the Telraam dataset. Similarly, as with the Madrid dataset, Interpolation and \ac{arima} display significant limitations for the Telraam data. Interpolation, in particular, shows notable increases in both proposed no ground truth and ground truth based metrics with growing gap sizes, indicating that it diverges from the original distribution. For instance, in the case of a 40-hour gap, interpolation results in a high \ac{jsd} and \ac{rmse}, which suggests that it fails to preserve the data’s distributional characteristics effectively. \ac{arima} demonstrates a similar trend with increasing \ac{jsd} and \ac{wd} for larger gaps, although it maintains slightly better stability than interpolation.  Additionally, the \ac{sarima} model shows moderate performance, achieving lower \ac{jsd} and \ac{wd} values than \ac{arima} and interpolation, though it still falls behind XGBoost and \ac{lstm}. However, for larger gaps, \ac{sarima} presents a higher divergence from the original data distribution, as seen in the proposed no ground truth and ground truth based metrics results. 

The analysis of both datasets reinforces the capabilities of \ac{jsd} and \ac{wd} as reliable no ground truth metrics for evaluating model performance in the absence of real values. The strong performance of XGBoost in the Telraam dataset and \ac{lstm} in the Madrid dataset highlights the potential of these models to adapt to dataset-specific characteristics. Meanwhile, the limitations of Interpolation and \ac{arima} across both datasets, as evidenced by higher \ac{jsd} and \ac{wd} values, further validate the performance of the proposed metrics in evaluating model quality. This alignment across models, datasets, and gap sizes underscores the suitability of \ac{jsd} and \ac{wd} as reliable performance metrics, enabling effective evaluation even in the absence of ground truth data.
\section{Conclusion}
\label{sec:conlusion}
This paper introduced \ac{wd} and \ac{jsd} as alternative validation metrics for evaluating data imputation techniques in the absence of ground truth data. By assessing the alignment between the distributions of imputed and pre-gap data, these metrics offer a reliable method for evaluating imputation quality based on statistical consistency rather than direct comparison with known values. Experimental results with Telraam and Madrid traffic datasets demonstrate that \ac{wd} and \ac{jsd} effectively capture imputation quality, suggesting their potential for broader applications in environments where ground truth is unavailable. Future work will explore integrating these metrics with adaptive ML models to further improve robustness and accuracy in complex data settings.

\section*{Acknowledgment}
The authors would like to thank Ericsson - Global Artificial Intelligence Accelerator (GAIA) AI-Hub Canada in Montréal and MITACS Accelerate under Grant IT38199, for funding and supporting this research.

\bibliographystyle{IEEEtran}
\bibliography{IEEEabrv,Biblio/references}

\begin{thebibliography}{10}
\providecommand{\url}[1]{#1}
\csname url@samestyle\endcsname
\providecommand{\newblock}{\relax}
\providecommand{\bibinfo}[2]{#2}
\providecommand{\BIBentrySTDinterwordspacing}{\spaceskip=0pt\relax}
\providecommand{\BIBentryALTinterwordstretchfactor}{4}
\providecommand{\BIBentryALTinterwordspacing}{\spaceskip=\fontdimen2\font plus
\BIBentryALTinterwordstretchfactor\fontdimen3\font minus \fontdimen4\font\relax}
\providecommand{\BIBforeignlanguage}[2]{{%
\expandafter\ifx\csname l@#1\endcsname\relax
\typeout{** WARNING: IEEEtran.bst: No hyphenation pattern has been}%
\typeout{** loaded for the language `#1'. Using the pattern for}%
\typeout{** the default language instead.}%
\else
\language=\csname l@#1\endcsname
\fi
#2}}
\providecommand{\BIBdecl}{\relax}
\BIBdecl

\bibitem{10327908}
B.~Jaumard and J.~M. Ziazet, ``5g e2e network slicing predictable traffic generator,'' in \emph{2023 19th Int. Conf. Network Service Manag. (CNSM)}, 2023, pp. 1--7.

\bibitem{9532728}
S.~Sarkar and A.~Debnath, ``Machine learning for 5g and beyond: Applications and future directions,'' in \emph{2021 Second Int. Conf. Electronics Sustainable Commun. Systems (ICESC)}, 2021, pp. 1688--1693.

\bibitem{9808164}
X.~Miao, Y.~Wu, L.~Chen, Y.~Gao, and J.~Yin, ``An experimental survey of missing data imputation algorithms,'' \emph{IEEE Trans. Knowl. Data Eng.}, vol.~35, no.~7, pp. 6630--6650, 2023.

\bibitem{DONDERS20061087}
\BIBentryALTinterwordspacing
A.~R.~T. Donders, G.~J. M.~G. van~der Heijden, T.~Stijnen, and K.~G.~M. Moons, ``Review: A gentle introduction to imputation of missing values,'' \emph{J. Clin. Epidemiol.}, vol.~59, no.~10, pp. 1087--1091, 2006. [Online]. Available: \url{https://www.sciencedirect.com/science/article/pii/S0895435606001971}
\BIBentrySTDinterwordspacing

\bibitem{JUNNINEN20042895}
\BIBentryALTinterwordspacing
H.~Junninen, H.~Niska, K.~Tuppurainen, J.~Ruuskanen, and M.~Kolehmainen, ``Methods for imputation of missing values in air quality data sets,'' \emph{Atmos. Environ.}, vol.~38, no.~18, pp. 2895--2907, 2004. [Online]. Available: \url{https://www.sciencedirect.com/science/article/pii/S1352231004001815}
\BIBentrySTDinterwordspacing

\bibitem{Tarsitano2011}
\BIBentryALTinterwordspacing
A.~Tarsitano and M.~Falcone, ``Missing-values adjustment for mixed-type data,'' \emph{J. Probab. Stat.}, vol. 2011, p. 290380, Aug 2011. [Online]. Available: \url{https://doi.org/10.1155/2011/290380}
\BIBentrySTDinterwordspacing

\bibitem{article2}
\BIBentryALTinterwordspacing
S.~Moritz, A.~Sardá-Espinosa, T.~Bartz-Beielstein, M.~Zaefferer, and J.~Stork, ``Comparison of different methods for univariate time series imputation in r,'' 2015. [Online]. Available: \url{https://arxiv.org/abs/1510.03924}
\BIBentrySTDinterwordspacing

\bibitem{article}
K.~Sutiene, G.~Vilutis, and D.~Sandonavicius, ``Forecasting of grid job waiting time from imputed time series,'' \emph{Electron. and Electr. Eng.}, vol. 114, Nov 2011.

\bibitem{atmos14020355}
\BIBentryALTinterwordspacing
L.~Wijesekara and L.~Liyanage, ``Mind the large gap: Novel algorithm using seasonal decomposition and elastic net regression to impute large intervals of missing data in air quality data,'' \emph{Atmosphere}, vol.~14, no.~2, 2023. [Online]. Available: \url{https://www.mdpi.com/2073-4433/14/2/355}
\BIBentrySTDinterwordspacing

\bibitem{Bandara_2021}
K.~Bandara, C.~Bergmeir, and H.~Hewamalage, ``{LSTM-MSNet}: Leveraging forecasts on sets of related time series with multiple seasonal patterns,'' \emph{IEEE Trans. Neural Networks Learn. Syst.}, vol.~32, no.~4, pp. 1586--1599, Apr 2021.

\bibitem{article3}
K.~Wellenzohn, M.~Böhlen, A.~Dignös, J.~Gamper, and H.~Mitterer, ``Continuous imputation of missing values in streams of pattern-determining time series,'' Mar 2017.

\bibitem{Stenger2024}
\BIBentryALTinterwordspacing
M.~Stenger, R.~Leppich, I.~Foster, S.~Kounev, and A.~Bauer, ``Evaluation is key: a survey on evaluation measures for synthetic time series,'' \emph{J. Big Data}, vol.~11, no.~1, p.~66, 2024. [Online]. Available: \url{https://doi.org/10.1186/s40537-024-00924-7}
\BIBentrySTDinterwordspacing

\bibitem{Villani2009}
\BIBentryALTinterwordspacing
C.~Villani, ``The wasserstein distances,'' in \emph{Optimal Transport: Old and New}.\hskip 1em plus 0.5em minus 0.4em\relax Berlin, Heidelberg: Springer Berlin Heidelberg, 2009, pp. 93--111. [Online]. Available: \url{https://doi.org/10.1007/978-3-540-71050-9_6}
\BIBentrySTDinterwordspacing

\bibitem{Telraam}
Telraam, ``Traffic count data,'' \url{https://telraam.net/}, jan 2024.

\bibitem{TelraamApi}
{City of Madrid}, ``Traffic data,'' \href{https://datos.madrid.es/portal/site/egob/menuitem.c05c1f754a33a9fbe4b2e4b284f1a5a0/?vgnextoid=33cb30c367e78410VgnVCM1000000b205a0aRCRD&vgnextchannel=374512b9ace9f310VgnVCM100000171f5a0aRCRD&vgnextfmt=default}{https://datos.madrid.es/portal/}, jan 2024.

\end{thebibliography}

\end{document}